\pdfoutput=1

\documentclass[11pt]{article}

\usepackage{EMNLP2023}

\usepackage{times}
\usepackage{latexsym}

\usepackage[T1]{fontenc}
\usepackage[utf8]{inputenc}

\usepackage{multirow}
\usepackage{microtype}
\usepackage{inconsolata}
\usepackage{fontawesome}
\usepackage{booktabs}
\usepackage{graphicx}
\usepackage{marvosym}

\title{Lil-Bevo: Explorations of Strategies for Training\\Language Models in More Humanlike Ways}

\author{Venkata S Govindarajan\Bat \quad Juan Diego Rodriguez\Faxmachine \quad Kaj Bostrom\Faxmachine \quad Kyle Mahowald\Bat  \\
\Bat Department of Linguistics \quad  \Faxmachine Department of Computer Science \\ The University of Texas at Austin\\
\texttt{\{venkatasg,juand-r,kaj,kyle\}@utexas.edu} \\
}

\begin{document}
\maketitle
\begin{abstract}
We present Lil-Bevo, our submission to the BabyLM Challenge. 
We pretrained our masked language models with three ingredients: an initial pretraining with music data, training on shorter sequences before training on longer ones, and masking specific tokens to target some of the BLiMP subtasks.
Overall, our baseline models performed above chance, but far below the performance levels of larger LLMs trained on more data.
We found that training on short sequences performed better than training on longer sequences.
Pretraining on music may help performance marginally, but, if so, the effect seems small.
Our targeted Masked Language Modeling augmentation did not seem to improve model performance in general, but did seem to help on some of the specific BLiMP tasks that we were targeting (e.g., Negative Polarity Items).
Training performant LLMs on small amounts of data is a difficult but potentially informative task.
While some of our techniques showed some promise, more work is needed to explore whether they can improve performance more than the modest gains here. Our code and models are available online.\footnote{\url{https://github.com/venkatasg/Lil-Bevo}}. 
\end{abstract}

\section{Introduction}\label{sec:introduction}

Large Language Models (LLMs) generate complex and largely grammatical strings and display impressive performance with structures traditionally thought to require abstract and hierarchical syntax \citep{linzen2016assessing,linzen2021syntactic,wilcoxLI,futrell2019rnns}. They have achieved human-like performance at a wide range of natural language tasks \citep{bubeck2023sparks,frank_2023}, particularly those having to do with linguistic \textit{form} \citep{mahowald2023dissociating}.
This state of affairs has led to claims that such models should be taken seriously as cognitive models of human language \citep{piantadosi2023modern,baroni2021proper,frank_2023}, in line with claims from the neuroscience literature to ``take mechanistic abstraction seriously'' \citep{cao2021explanatory}.

One reason that has been posited \textit{not} to take LLMs seriously as cognitive models, though, is the immense amount of data they are trained on relative to what a human child is exposed to \citep{warstadt2022what,van-schijndel-etal-2019-quantity}.
Thus, it is possible that models memorize more than humans do and, relative to humans, over-rely on statistical heuristics and memorized chunks of language \citep{bender2021dangers}.

On the other hand, the quality of data that LLMs get during pretraining is, in many ways, much worse than what human learners get. 
Children get richly structured, interactive, multimodal input, tailored to their specific interests and needs.
A baby might reach for a cup of water and be told ``Water. You want some water?'' 
Given that babies are known to conduct repeated experiments to learn about the world \citep{gopnik1999scientist}, the baby might try this again and again until mastering the concept of what water is.
An LLM, meanwhile, might begin learning language by being asked to predict random tokens in the Wikipedia article on quantum mechanics.

In this paper, we describe our experiments with Lil-Bevo, a small language model trained on human-scale data for the BabyLM competition \citep{warstadt-et-al-2023-babylm}.
The goal of the competition is to train a performant LM on a human-scale amount of data: 10M words for the small track, 100M for the larger track.
We submitted to both strict tracks --- however, we were notified through the meta-review that our models qualify only for the loose track due to the usage of additional non-linguistic data (music from the MAESTRO dataset~\citep{hawthorne2018enabling}).
The evaluation is on a set of natural language tasks including grammatical acceptability judgments via minimal pairs in the BLiMP benchmark \citep{warstadt2020blimp}, language understanding tasks in SuperGLUE \citep{wang2019superglue}, and MSGS (the Mixed Signals Generalization Set) \citep{warstadt-etal-2020-learning}

We started with a baseline DeBERTa model, trained from scratch on BabyLM data using a custom unigram SentencePiece tokenizer ~\citep{kudo-richardson-2018-sentencepiece}.
Our strategy was not focused on the architecture, but on ways in which we could adjust the training regime to improve performance above the baseline.

Specifically, our strategy targets 3 ways in which typical LLM training regimes lead to lower-quality data than humans have access to.
Here, we describe those strategies and their motivation. 
We give detailed methods in Section~\ref{sec:experiments_and_methods} and then present results, including a number of ablation studies that attempt to partition out what strategies were successful.

We treated these studies as proof-of-concept and did not exhaustively test these strategies.
Thus, we think that there is still room for improvement.

\paragraph{Training on Short Sequences}

Unlike LLMs, babies do not start language by learning long complicated sequences all at once.
Using databases of child and child-directed speech, it has been shown that there is some alignment of caretakers to the child's level in terms of linguistic complexity such that caregivers talk to younger children using shorter utterances and longer utterances as they develop \citep{schwab2016language,kunert2011adaptation}. 
To that end, \citet{mueller-linzen-2023-plant} showed that training on simpler data first could induce a better hierarchical bias for learning language.
We specifically take inspiration from \citet{press-etal-2021-shortformer} who showed that LLMs learn better when trained on shorter sequences before being trained on longer sequences.

\paragraph{Training on Music Before Training on Language}
Unlike LLMs, babies are exposed to a wide range of input besides just text.
Before and while learning language, they are also learning to map the visual world, to navigate the physical world, to process non-linguistic auditory stimuli, and to engage in a wide variety of cognitive operations.
Thus, it is commonly observed that some of the machinery thought to be language-specific (e.g., hierarchical structure) might be induced in pre-linguistic infants through exposure to other kinds of stimuli.
\citet{papadimitriou-jurafsky-2020-learning} use this idea to show that training language models on structured data (e.g., music) can help models learn faster.
We use a similar idea, with initial pretraining on a mix of music (piano performances) and text.

\paragraph{Targeted Masked Language Model} The role of child-directed speech in human language learning is controversial \citep[see][for discussion and a large-scale replication of infant-directed speech preferences]{manybabies2020quantifying}.
It is generally agreed that parents do not correct a child every time they make a grammatical error \citep{marcus1993negative}, but there is also evidence that social feedback acts as a signal \citep{tomasello1992social} and that parents structure input to be helpful \citep{weisleder2013talking}.
When a child says something wrong, a parent might ``recast'' the utterance or highlight grammatical features that children are struggling with \citep{nicholas2001recasts}. 
Inspired by this idea, targeting the BLiMP \citep{warstadt2020blimp} syntactic evaluations as well as more general tasks, we trained with a targeted MLM objective. 

We considered some variations of the idea of learning with some external feedback that distinguishes correct tokens against corrupted/noisy replacements. For example, ELECTRA \citep{Clark2020ELECTRA} consists in learning to detect tokens which have been replaced by an auxiliary model. Unfortunately, replaced token detection approaches such as ELECTRA \citep{Clark2020ELECTRA} suffer from an inability to learn probability distributions over the entire vocabulary, and so cannot be used for (pseudo)-likelihood scoring \cite{salazar-etal-2020-masked}.  Another related approach is Corrective Language Modeling (CLM) \citep{DBLP:journals/corr/abs-2204-06644}, in which the model is trained to correctly replace corrupted tokens; however, it is not clear how to best use these models for scoring sentences in BLiMP.\footnote{Initial experiments with CLM performed worse than masked language modeling (MLM); we believe this is due to a mismatch between training and how the pseudo-likelihood scoring is done via masking.}

Given the problems outlined above, we decided to use masked language modeling (MLM) with targeted masks. The motivation is to make it easier for the model to learn syntactic phenomena that co-occur frequently with certain words. Other strategies for selecting masks were used in \citet{sadeq-etal-2022-informask,gu-etal-2020-train}; unlike these works, we mask specific words which are essential to the phenomena in BLiMP. For example, to target the filler-gap dependency subtask in BLiMP, we go through the original data set and mask every occurrence of ``that'' and ``what'' in the corpus. By focusing on these words, we anticipate that the model will more quickly learn to score ``I know what you did last summer.'' more highly than ``I know that you did last summer.''

\section{Experiments \& Methods} \label{sec:experiments_and_methods}

We report all experiments and results for Lil-Bevo in this paper, as it enabled quick prototyping, and because we find similar trends with our larger model Lil-Bevo-X. Lil-Bevo-X differs from Lil-Bevo in the model used (\texttt{deberta-base} rather than \texttt{deberta-small}), training data (100M versus 10M), and vocabulary size. Final results for the Lil-Bevo-X are available on our \href{https://github.com/venkatasg/Lil-Bevo}{online repository}.

\paragraph{Tokenizer} We trained a unigram SentencePiece tokenizer~\citep{kudo-richardson-2018-sentencepiece} from scratch on the BabyLM data combined with the MAESTRO~\citep{hawthorne2018enabling} dataset (described in detail below)  using the \texttt{sentencepiece} library. Specifically, we trained a tokenizer with a vocabulary size of 16,640 and 33,280 for Lil-Bevo and Lil-Bevo-X respectively. \texttt{<mask>} and \texttt{<cls>} were included as control symbols in the vocabulary, along with an end-of-sequence token (\texttt{</s>}), a pad token (\texttt{<pad>}) and an unknown token (\texttt{<unk>}).

\paragraph{Model} We chose to use an encoder-based language model, specifically DeBERTa since (a) encoder-based language models are known to capture many syntactic and semantic features in language when pretrained on relatively modest amounts of data 
\citep{zhang-etal-2021-need}, (b) there were a wide variety of off-the-shelf DeBERTa architectures available on HuggingFace for easy prototyping and use. 

We trained the model in three phrases: (1) pretraining on a combination of music and text for \textbf{5} epochs with a sequence length of 64 tokens, (2) continuing pretraining on text for \textbf{50} epochs with a sequence length of 128 tokens, and (3) finally pretraining on text using targeted MLM for \textbf{2} epochs with a sequence length of 512 tokens. Each of these is described in more detail below.  

\paragraph{1. Music Pretraining}\label{para:music} \citet{papadimitriou-jurafsky-2020-learning} find that pretraining on languages other than the target language --- including music and code --- lead to lower perplexities on target language as compared to random distributions of tokens, or even Zipfian token distributions. Inspired by this idea, we explored whether supplementing the 10M linguistic tokens with \emph{non-linguistic} musical tokens from the MAESTRO dataset~\citep{hawthorne2018enabling} could lead to noticeable improvements in LM learning. The impetus behind pretraining on music is two-fold: (a) additional training data that nevertheless has structural biases that could help the model learn structural biases found in language (b) the model reaching a stable region in parameter space that enables it to learn desired linguistic properties much faster and/or better.

After several experiments, we found that pretraining on the combined \emph{strict-small} and the entire MAESTRO dataset for 5 epochs provided the best results. We use V3.0.0 of the MAESTRO dataset, which contains 85M tokens using our custom trained tokenizer. The dataset consists of 200 hours of MIDI piano recordings, which we convert to text and tokenize with the shared unigram SentencePiece tokenizer. Our textual representation of MIDI consists of a chronological sequence of codes describing the channel and key of each note onset and release event (e.g. \texttt{c0n71} for `note on, channel 0, key 71') delimited by spaces and optional codes for time between events (e.g. \texttt{t18} for 18 MIDI ticks).  We chose a short sequence length of 64 tokens for pretraining inspired by the Shortformer, which we now explain in further detail.

\paragraph{2. Shortformer} \citet{press-etal-2021-shortformer} introduce a few innovations to the training regime. 
In particular, we focused on their idea of training for shorter sequence lengths before moving onto longer ones.
We used a similar training regime to \citep{press-etal-2021-shortformer}, where we started with a training sequence length of 128 for 50 epochs, before moving to a training sequence length of 512. We initially experimented with training on longer subsequence length for 150 epochs as in \citet{press-etal-2021-shortformer}, but discovered lower evaluation results on most BLiMP categories (albeit with some improvements on some categories like Island Effects and Quantifiers). Results on BLiMP \citep{warstadt2020blimp} and SuperGLUE \citep{wang2019superglue} saturated with as little as 2 epochs --- we believe this is because of the much smaller size of the dataset as compared to that in \citep{press-etal-2021-shortformer}, leading to overfitting on the dataset.

\paragraph{3. Targeted MLM}

We specifically masked out words which were essential to some of the BLIMP subtasks. Some of these, such as quantifier and negation words, are also important to some of the SuperGLUE tasks (e.g., textual entailment.) For anaphor agreement, we masked the words ``himself'', ``herself'', ``itself'', ``themselves''. For NPI licensing the masked words included ``not'', ``often'', and ``probably''\footnote{Note that the masked words are not necessarily NPI items themselves, but rather that they are targets of single word substitutions in NPI items.}.  The list of words which were masked in each category are shown in Table~\ref{tab:mask_stats} in Appendix~\ref{sec:appendix}. We used a sequence length of 512 tokens, and additionally masked other random tokens in order to mask a total of 15\% of tokens per sample. 

The total number of words masked for each category across the 10M train set are given in Table~\ref{table:substitution_stats}. 

\begin{table}[t]
\centering
\begin{tabular}{lrr}
\toprule
\textbf{Category} & \textbf{Total} & \textbf{Avg} \\ 
\midrule
S-V agreement  & 124197 & 4.3 \\
Animacy        & 100206 & 3.5 \\
Quantifiers    & 89926  & 3.1 \\
Modal verbs    & 58604  & 2.0 \\
NPI licensing  & 47484  & 1.6 \\ 
Filler gap     & 34988  & 1.2 \\
D-N agreement  & 28675  & 1.0 \\
Adverbs        & 19332  & 0.7 \\
Anaphor agreement & 3659 & 0.1\\
\bottomrule
\end{tabular}
\caption{Total number of masks and average number of masks per sample for each targeted category (\emph{S-V agreement} stands for subject-verb agreement, and \emph{D-N agreement} stands for determiner-noun agreement).}
\label{table:substitution_stats}
\end{table}

The \emph{Animacy} class consists of animate nouns, and was used to target the minimal pairs in the \emph{Argument Structure} category with animate/inanimate subjects (``Amanda was respected by some \emph{waitresses}.'' vs ``Amanda was respected by some picture''). To obtain a list of animate nouns we used all the lemmas of (direct and indirect) hyponym synsets of \emph{person.n.01} in WordNet.

In addition to targeting the BLiMP categories of S-V agreement, quantifiers, NPI licensing, filler gap, argument structure, DN- agreement and anaphor agreement, we also included some \emph{modal verbs} (e.g., can, might, shall) and certain \emph{adverbs} (e.g., never, maybe, always, perhaps), since these are important for textual entailment.

\subsection{Ablations}

We compare Lil-Bevo with ablations to explore how important our three strategies are for final performance. Specifically, we compare Lil-Bevo with the following:

\paragraph{Long-only} Train DeBERTa with a sequence length of 512 tokens for 57 epochs.

\paragraph{Short-only} Train DeBERTa with a sequence length of 128 tokens for 57 epochs.

\paragraph{Short+target} Train DeBERTa with a sequence length of 128 tokens for 55 epochs. Then train with targeted MLM for 2 epochs.

\paragraph{Music+short} Train DeBERTa on music and text for 5 epochs with a sequence length of 64 tokens. Then continue training on text with a sequence length of 128 tokens for 52 epochs.

\paragraph{Music+short+long} Train DeBERTa on music and text for 5 epochs with a sequence length of 64 tokens. Then continue training on text with a sequence length of 128 tokens for 50 epochs, followed by training with a sequence length of 512 tokens for 2 epochs.

\paragraph{Lil-Bevo (music+short±target)} This is the same as \emph{Music+short+long} except that the final stage of pretraining for 2 epochs uses targeted MLM.

\paragraph{Implementation} We train all our models using the \texttt{Trainer} API, part of the \texttt{huggingface} python package. Models are trained using 4 Nvidia A40 GPUs, with the maximum possible batch size that was permissible with each experiment. Apart from setting initial learning rate to 6e-4, weight decay to 0.1 and a warmup ratio to 0.0001, we use default training arguments in the API (except for the final targeted MLM/long stage, where we used all default parameters). Models are evaluated on the validation split of the BabyLM dataset. We did not use the test split of the BabyLM data. We release all of the above pretrained models \href{https://huggingface.co/collections/venkatasg/babylm-653591cdb66f4bf68922873a}{online on the Huggingface Hub}.

\section{Results}  \label{sec:results}

\begin{figure*}
    \centering
    \includegraphics[width=\linewidth]{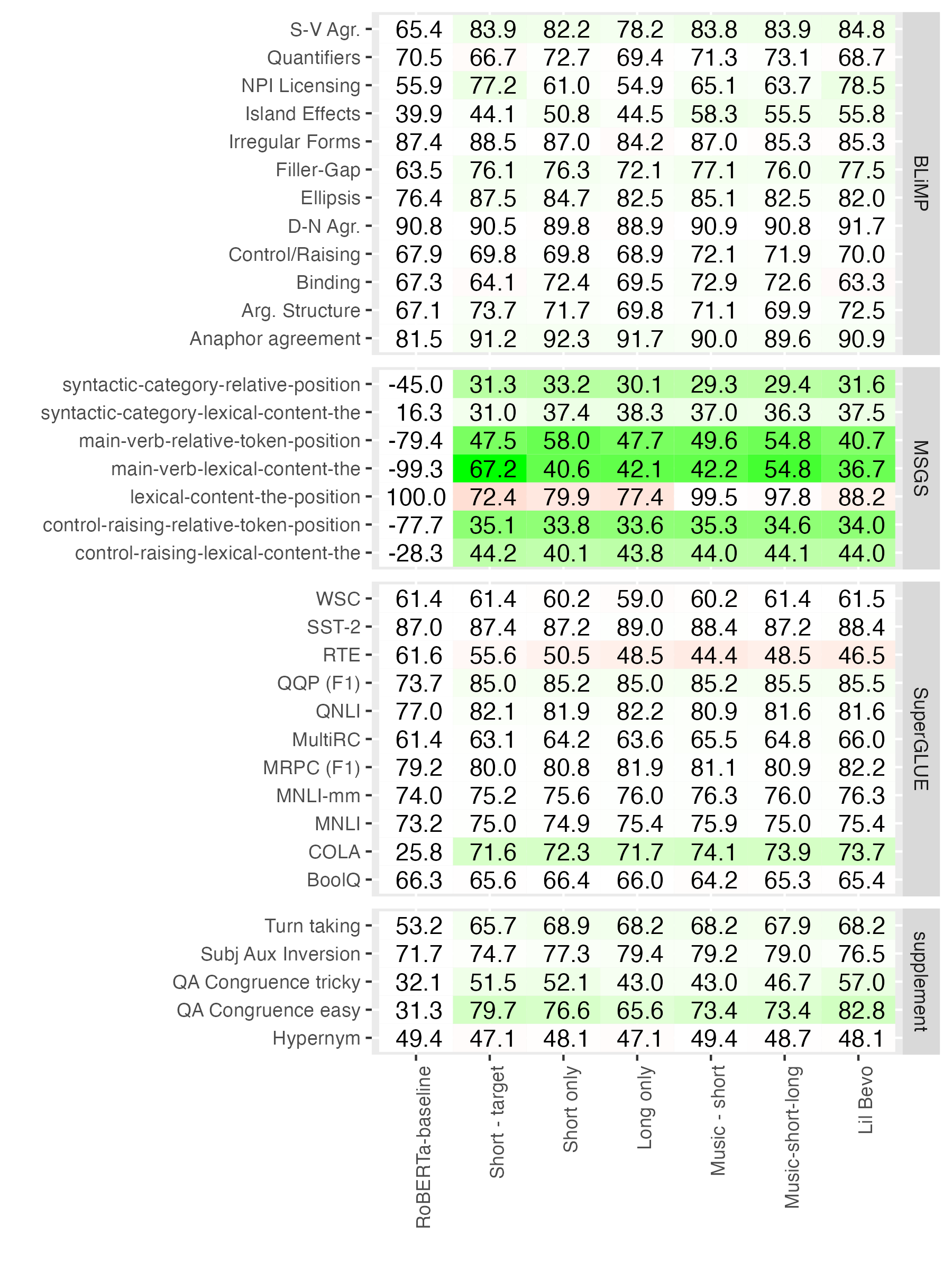}
    \caption{Results for each model, for each task. The color reflects the difference in score between the given model and the RoBERTa baseline results released by the organizers of BabyLM.}
    \label{fig:enter-label}
\end{figure*}

Results for BLiMP, MSGS, SuperGLUE and the supplementary tasks are shown in Figure~\ref{fig:enter-label}.  The results are color-coded to represent each model's differences from the RoBERTa baseline results (obtained from the BabyLM GitHub). We highlight some results below.

\paragraph{Does pretraining on music help?} Comparing \emph{short-only} with \emph{music+short}, we see that pretraining on music helps slightly on 8 of the 12 BLiMP subtasks, and on two of the 5 supplement tasks. However, it suffers from a large gap of 9.1 points on \emph{QA Congruence tricky}. On SuperGLUE, \emph{music+short} outperforms \emph{short-only} on 6 of the 11 subtasks, and only slightly. 
Thus, we do not think there is strong evidence that pretraining on music improves over the short-only condition, in isolation.

Comparing \emph{Lil-Bevo} (music+short+target) with \emph{short+target}, we see that \emph{Lil-Bevo} outperforms \emph{short+target} on 69\% of all tasks. 
Predicting score for each task in a mixed-effect linear regression with a fixed effect predictor for whether the model was \emph{Lil-Bevo} or \emph{short+target}, we found that \emph{Lil-Bevo} was slightly better ($\beta = 1.3$, $\chi^2(1) = 4.11, p < .05$ by a likelihood ratio test).
So, while music pretraining may help, the effect is small and inconsistent in our observed data.

\paragraph{What is the effect of targeted MLM?} We compare \emph{music+short+long} with Lil-Bevo (music+short+target) and \emph{short-only} with \emph{short+target} to ascertain whether targeted MLM helps over random masking. Targeted MLM does not systematically improve performance, except for two BLiMP tasks: NPI Licensing and Argument Structure. For NPI Licensing, Lil-Bevo outperforms \emph{music+short+long} by 14.8 points, and \emph{short+target} outperforms \emph{short-only} by 16.2 points.
We suspect that this difference could be meaningful since our Targeted MLM strategy specifically targets NPI terms that are substituted in BLiMP.

\paragraph{The effect of increasing sequence length} When comparing \emph{music+short} with \emph{music-short-long}, and \emph{short-only} with \emph{long-only}, we find that pretraining with 512-token sequence lengths generally underperforms pretraining with 128-token sequence lengths.
The difference between \emph{short-only} and \emph{long-only} conditions is quite large in fact.
A linear mixed effect regression comparing the two using the same method as above found that performance was 1.8 points worse on average for the \emph{long-only} method ($\beta = 1.8$, $\chi^2(1) = 14.2, p < .001$ by a likelihood ratio test).
Thus, we believe pretraining with shorter sequences helps significantly compared to using longer sequences.

\section{Discussion} \label{sec:analysis}

Overall, we found that, for BabyLM's, sequence length matters, music pretraining may help a little (but may be spurious), and targeted MLM training may help on specific tasks.

These results are far from exhaustive, and we see a number of areas for future improvement using these methods.
To fully understand the role of initial pretraining on music, one could construct a series of synthetically-generated music datasets, with varying degrees of complexity. Would pretraining on music that is more ``language-like'' \citep{lerdahl1996generative} in some sense improve performance on downstream tasks? Perhaps there is a principled way to interpolate between music and language, using the same kind of data format (MIDI). At one end of the spectrum one would have MAESTRO, and at the other end, text that has been encoded into MIDI events. 

Related to the use of varying sequence lengths, future work could consider improvements in data preprocessing and batching; in particular, knowing the beginning and ending of coherent chunks of text (e.g., dialogues or documents) could help improve the model. Beyond this, \citet{mueller-linzen-2023-plant} provide some evidence that curriculum learning approaches may be fruitful to improving low-resource language models.

Finally, a more thorough analysis is needed on when (and by how much) targeted MLM is able to boost model performance. Other strategies are also possible, such as combining targeted MLM with information-theoretic strategies for picking random masks \citep{sadeq-etal-2022-informask}. Beyond MLM, contrastive objectives could be used to encourage the model to score grammatical sentences more highly than ungrammatical sentences.

\begin{table}[]
    \centering
    \begin{tabular}{ll}
        \toprule
        \textbf{Model} & \textbf{Dynabench score} \\\midrule
        Lil-Bevo & 0.64 \\
        Music-short-long & 0.64 \\
        Music-short & 0.69 \\
        Short-only & 0.63 \\
        Short-target & 0.62 \\
        Long-only & 0.61 \\
        Lil-Bevo-X & 0.69 \\
        \bottomrule
    \end{tabular}
    \caption{Scores on Dynabench for different models.}
    \label{tab:dynabench}
\end{table}

\section{Conclusion} \label{sec:conclusion}

A big motivating question for training models on human-scale data is whether it is possible for models to attain linguistic competence without the massive amounts of data used to train the massive LLMs that dominate NLP leaderboards.
If so, that would make it more plausible that we should take LLMs seriously as cognitive models.
So can BabyLMs learn like grown-up ones?
While we find some hints of directions to pursue for making small language models learn more from less, we did not come close to matching LLM performance from larger amounts of data. 
Of course, that does not mean it is not possible to do so, and other teams might have different experiences. 
We did not fully explore optimizing all of our methods, and we treated our manipulations largely as proof-of-concept.
Aggregating methods and results from a wider variety of teams will make it possible to more fully explore these questions.

\bibliography{references}
\bibliographystyle{acl_natbib}

\appendix

\section{Appendix}
\label{sec:appendix}

Table~\ref{tab:mask_stats} shows the list of words selected for targeted MLM for each linguistic category, while age of acquisition results are presented in Table~\ref{tab:aoa}

\begin{table*}[!h]
\centering
\begin{tabular}{ll}
\toprule
\textbf{Category} & \textbf{Words} \\
\multirow{2}{*}{S-V agreement} & is, was, have, do, are, don't, were, has, does, isn't, doesn't, wasn't, haven't, \\
 & aren't, weren't, hasn't\\
 \midrule
\multirow{2}{*}{Quantifiers} & all, some, more, any, little, many, much, most, every, both, each, few, enough,  \\
 & several,half, less,either, none,. lots, neither, plenty \\
 \midrule
Filler gap & that  \\
 \midrule
Modal verbs & can, would, will, could, should, may, must, might, shall \\
\midrule
NPI licensing & not, only, also, really, probably, often, certainly, clearly \\
 \midrule
D-N agreement & this, these \\
 \midrule
\multirow{3}{*}{Adverbs} & never, always, maybe, probably, perhaps, certainly, absolutely, likely, possibly, \\
 & definitely, surely, truly, constantly, forever, potentially, positively, undoubtedly,  \\
 & consistently, invariably, eternally, perpetually, dubiously, uncertainly\\
 \midrule
 Anaphor agreement & himself, themselves, itself, herself \\
 \midrule
 \multirow{3}{*}{Animacy} & people, man, men, family, person, father, mother, girl, woman, son, children, \\
  & guy, friend, wife, boy, guys, human, member, friends, women, members, \\
  & daughter, child, brother, boys, husband, girls, lady, parents, kids, king, sister, dad,\\
  & mommy, daddy, player, students, doctor, president, captain, kid, mom, leader,\\
  & officer, director, players, soldiers, teacher, god, student, sir, officers, judge, patient,\\
  & brothers, families, mark, actor, ladies, singer, uncle, author, manager, gentleman,\\
  & humans, lad, writer, sweetie, prince, lawyer, artist, mum, host, owner, guest,\\
  & teachers, princess, scientists, guard, professor, artists, leaders, agent, assistant,\\ & patients, mama, workers, minister, boss, sons, criminal, partner, babies, citizens,\\
  & adult, politician, gods, mayor, actress, principal, cousin, witness, driver, hero,\\
  & governor, lord, doctors, authorities, maiden, suspect, victims, aunt, candidate,\\
  & individuals, producer, champion, gentlemen, founder, enemies, sisters, winner,\\
  & passenger, client, bride, priest, prisoners, pilot, inhabitants, ghost, chairman,\\
  & nurse, guests, user, pirate, graduate, merchant, cats, victim, passengers, pirates,\\
  & noble, agents, expert, parent, editor, grandma, officials, subjects, cops, maid,\\
  & commander, policeman, writers, servants, academic, peasant, eldest, engineer,\\ 
  & musician, devil, critics, users, creatures, twin, composer, personality, lads,\\
  & followers, poet, adults, boyfriend, fellows, actors, ruler, judges, witch, daughters,\\
  & lieutenant, musicians, servant, secretary, slave, priests, scholars, prisoner,\\
  & visitors, residents, lover, cop, companion, knight, deputy, customers, tourist,\\
  & guards, grandfather, journalist, architect, rival, kings, colleagues, farmers,\\
  & owners, farmer,...\\
 \bottomrule
\end{tabular}
\caption{Words which were masked in targeted MLM in the 10M train set. For \emph{Animacy} only words appearing over 100 times are shown in the table.}
\label{tab:mask_stats}
\end{table*}

\begin{table*}[h]
    \centering
    \begin{tabular}{lllll}
        \toprule
        Model & Overall &	Nouns &	Predicates &	Function words \\\midrule
         RoBERTa-baseline & 2.06 & 1.99 & 1.85 & 2.65  \\
         Lil-Bevo & 2.06 & 2.0 & 1.84 & 2.65  \\
         Lil-Bevo-X & 2.05 & 1.99 & 1.85 & 2.59 \\
         \bottomrule
    \end{tabular}
    \caption{Age of Acquisiton results}
    \label{tab:aoa}
\end{table*}

\end{document}